\documentclass[a4paper,fleqn]{cas-sc}

\usepackage[authoryear,longnamesfirst]{natbib}
\setcitestyle{numbers}

\def\tsc#1{\csdef{#1}{\textsc{\lowercase{#1}}\xspace}}
\tsc{WGM}
\tsc{QE}
\begin{document}
\let\WriteBookmarks\relax
\def\floatpagepagefraction{1}
\def\textpagefraction{.001}

\shorttitle{MF-LPR²}    

\shortauthors{Na et al.}

\title [mode = title]{MF-LPR²: \textnormal{Multi-Frame License Plate Image Restoration and Recognition using Optical Flow}}

\tnotemark[1]
\tnotetext[1]{Our RLPR dataset can be found in 10.17632/4rs5wpvckz.2}

\author[1]{Kihyun Na}[orcid=0009-0001-3827-5371]
% % Footnote of the first author
% % Footnote text
\ead{kevinna95@gmail.com}
\credit{Conceptualization of this study, Methodology, Software, Writing, Validation, Supervision}

% Author
\author[1]{Junseok Oh}[orcid=0009-0001-5450-8956]
\ead{junseokoh96@gmail.com}
\credit{Conceptualization of this study, Methodology, Software}
\fnmark[1]
\fntext[1]{Currently, working at Samsung Electronics.}

\author[1]{Youngkwan Cho}[orcid=0009-0004-9821-8688]
\ead{dudrhks1009@naver.com}
\credit{Software, Data Curation, Validation, Writing - Original Draft}

\author[1]{Bumjin Kim}[orcid=0009-0009-6331-6938]
\ead{21900104@handong.ac.kr}
\credit{Software, Data Curation, Validation, Visualization}

\author[1]{Sungmin Cho}[orcid=0009-0008-0815-5441]
\ead{ko041213@gmail.com}
\credit{Data Curation}

\author[1]{Jinyoung Choi}[orcid=0009-0002-6255-2882]
\ead{jinyoung@handong.ac.kr}
\credit{Validation, Writing - Review \& Editing}

\author[1]{Injung Kim}[orcid=0000-0003-4439-6097]
\ead{ijkim@handong.edu}
\credit{Conceptualization of this study, Methodology, Writing - Review \& Editing, Project administration}
\cormark[1]
\cortext[1]{Corresponding author}

\affiliation[1]{organization={Department of Computer Science and Electrical Engineering (CSEE), Handong Global University},
            addressline={558 Handong-ro, Buk-gu}, 
            city={Pohang},
            postcode={37554}, 
            state={Gyeongsangbuk},
            country={Republic of Korea}}


\begin{abstract}
License plate recognition (LPR) is important for traffic law enforcement, crime investigation, and surveillance. However, license plate areas in dash cam images often suffer from low resolution, motion blur, and glare, which make accurate recognition challenging. Existing generative models that rely on pretrained priors cannot reliably restore such poor-quality images, frequently introducing severe artifacts and distortions. To address this issue, we propose a novel multi-frame license plate restoration and recognition framework, MF-LPR², which addresses ambiguities in poor-quality images by aligning and aggregating neighboring frames instead of relying on pretrained knowledge.
To achieve accurate frame alignment, we employ a state-of-the-art optical flow estimator in conjunction with carefully designed algorithms that detect and correct erroneous optical flow estimations by leveraging the spatio-temporal consistency inherent in license plate image sequences. Our approach enhances both image quality and recognition accuracy while preserving the evidential content of the input images. 
In addition, we constructed a novel Realistic LPR (RLPR) dataset to evaluate MF-LPR². The RLPR dataset contains 200 pairs of low-quality license plate image sequences and high-quality pseudo ground-truth images, reflecting the complexities of real-world scenarios. In experiments, MF-LPR² outperformed eight recent restoration models in terms of PSNR, SSIM, and LPIPS by significant margins. In recognition, MF-LPR² achieved an accuracy of 86.44\%, outperforming both the best single-frame LPR (14.04\%) and the multi-frame LPR (82.55\%) among the eleven baseline models. The results of ablation studies confirm that our filtering and refinement algorithms significantly contribute to these improvements.
\end{abstract} 

% %% (Optional): https://www.elsevier.com/researcher/author/tools-and-resources/graphical-abstract 참고
% \includegraphics[width=6.5in]{MFLPR_Framework.pdf}
% \includegraphics[width=6.5in]{RLPR_Framework.pdf}

% \begin{highlights} %% (Mandatory): https://www.elsevier.com/researcher/author/tools-and-resources/highlights 참고
% \item Propose MF-LPR², a multi-frame framework to restore low-quality license plates.
% \item Align frames via novel optical flow filtering and refinement for robust restoration.
% \item Introduces PDNF‑k metric to measure spurious artifacts in restoration outputs.
% \item Presents RLPR dataset: 200 dash cam sequences with 31 frames each for testing.
% \item Achieve notable improvement over baselines in image quality and recognition accuracy.
% \end{highlights}

% Keywords

\maketitle
% --- AAM 고지(타이틀 아래 박스) ---
\begin{center}
\begin{minipage}{0.96\linewidth}
\small
\textbf{Author Accepted Manuscript (AAM).} 
\copyright~2025.
This manuscript version is made available under the \textbf{CC BY-NC-ND 4.0} license (\url{https://creativecommons.org/licenses/by-nc-nd/4.0/}).
Please cite the published article in \emph{Computer Vision and Image Understanding} (2025). 
DOI: \href{https://doi.org/10.1016/j.cviu.2025.104361}{10.1016/j.cviu.2025.104361}.
\end{minipage}
\end{center}

% Main text
\section{Introduction}\label{Introduction}
License Plate Recognition (LPR) plays a crucial role in various traffic applications, including traffic law enforcement, crime investigation, and surveillance. However, in road video footage, license plates often appear small, even in high-resolution images (e.g., 2560x1440). As a result, the resolutions of the license plate image are often insufficient for recognition (e.g., 88x32). This issue is further compounded by other degradations, such as motion blur, light glare, and inaccurate focus, which makes reliable recognition challenging in real-world scenarios.

License plate image restoration is distinguished from general image restoration, which primarily focuses on visual quality, in that it must enhance legibility while preserving the evidential value of the original footage.
For example, the model must not restore the degraded image of the digit `6' in a way that makes it resemble `5' or `8', as such errors compromise the forensic value of the image.

However, most recent restoration methods often fail to preserve the original information of images, introducing artifacts or distortions due to their heavy reliance on prior knowledge acquired from training data.

\begin{figure}
\centering
\includegraphics[width=6.5in]{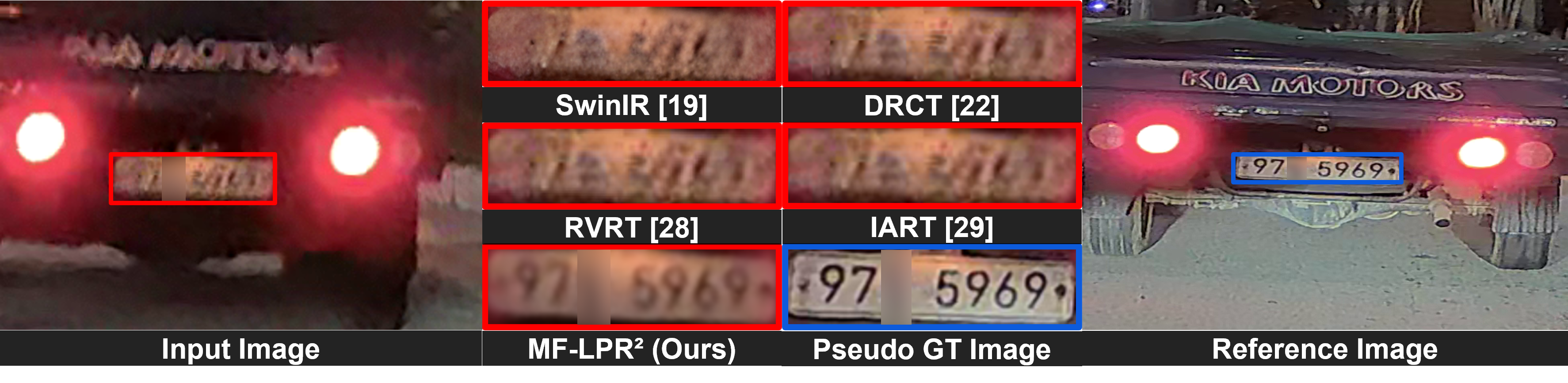}
\caption{ The result of the proposed model (MF-LPR²) compared with two single image super resolution (SISR), SwinIR\cite{SwinIR}, DRCT\cite{DRCT}, and two video restoration models (VSR), RVRT \cite{RVRT} and IART \cite{IART}. The left column displays the low-quality input image. The middle six images present the results of the five restoration models and as well as the pseudo ground-truth image extracted from the reference image (right column). MF-LPR² showed the best result both preserving the content of the input image and improving visibility.}
\label{fig:scenario}
\end{figure}

The advent of deep generative models has led to significant advancements in image restoration, enabling the reconstruction of missing details leveraging pretrained knowledge acquired from large datasets. This approach has been widely applied in tasks such as single-image super-resolution (SISR) \cite{SRCNN, SRGAN, ESRGAN, EDSR, RCAN, HAT, Real-ESRGAN, BSRGAN, IPT, SwinIR, Restormer, Uformer, DRCT, DCLS}, multi-frame super-resolution (MFSR) \cite{EyesOnTarget}, and video super-resolution (VSR) \cite{BasicVSR++, PSRT-recurrent, EDVR, VRT, RVRT, IART}, demonstrating substantial improvements in visual quality. In particular, recently developed scene text image super-resolution (STISR) models \cite{TextSR, TextZoom, TPGSR, STT, TATT, DPMN} have shown their effectiveness in enhancing the resolution and readability of low-resolution text images by leveraging the semantic information unique to textual contents.

However, existing generative models often struggle with severely degraded license plate images, as shown in Fig. \ref{fig:scenario}. This challenge is particularly pronounced for images that differ significantly from the characteristics of the training data. The primary reasons for this limitation is as follows: First, the degradation in the input image quality surpasses the range that restoration models can handle. Second, generative models tend to prioritize the enhancement of visual quality by relying heavily on pretrained knowledge, rather than focusing on preserving the content of the input images. 
Third, restoration models are generally trained on synthetic image pairs consisting of high-resolution images and their artificially degraded counterparts, as collecting paired datasets of low- and high-quality license plate images is resource-intensive and costly. However, the degradations observed in real-world license plate images differ significantly from these artificial degradations.

There are several previous studies on license plate image restoration \cite{svoboda2016cnn, lu2016robust, rao2017new, nguyen2018joint, EyesOnTarget}. Most of these studies focus on motion deblurring and the issues related to low resolution and quality have received relatively less attention.
While several previous studies have attempted to extend restoration beyond motion deblurring \cite{SemanticVLPSR, LPR-RSR-EXT, AFANet}, these models were trained on synthetic datasets, limiting their effectiveness in real-world scenarios with diverse degradation patterns.
 
In this study, our objective is to restore low-quality license plate images in a manner that enhances their legibility while preserving the evidential value by preventing any alteration of the original content. To achieve this goal, we propose a novel multi-frame license plate image restoration and recognition (MF-LPR²) framework. Unlike generative models that rely on prior knowledge obtained from training data, MF-LPR² restores high-quality license plate images by combining the complementary information from multiple low-resolution input frames.

MF-LPR² aligns a sequence of low-quality frames using optical flows and aggregates the aligned frames to produce a high-quality output image. However, while accurate optical flow estimation is critical in this approach, even state-of-the-art optical flow estimators often produce erroneous results for low-quality images, which severely degrades output quality as Fig. \ref{fig:Spatial_diff_image}. To overcome this challenge, we present reliable optical flow filtering and refinement algorithms based on spatio-temporal consistency inherent in license plate image sequences. The experimental results in Subsection \ref{subsec:filtering_and_refinement} demonstrates the effectiveness of the proposed filtering and refinement algorithms. Especially, Table \ref{table:module_effectiveness} presents the results of ablation studies showing that our algorithms increase the character recognition accuracy by 11.74\%p from 74.70\% to 86.44\%.

Leveraging the precisely aligned neighboring frames, MF-LPR² effectively restores severely degraded license plate images, as shown in Fig. \ref{fig:scenario}. The most significant advantage of MF-LPR² over existing generative models is its ability to preserve the content of the input image by avoiding the generation of spurious artifacts and distortions. This feature is essential for maintaining the evidential value of the original dash cam footage. Therefore, we propose a novel metric, the top-$k$ Percentile average Distance to Nearest Frame (PDNF-$k$), to quantify the severity of local artifacts in the restored image.

In addition, we constructed the Realistic License Plate Restoration and Recognition (RLPR) dataset to evaluate the proposed framework. The RLPR dataset contains 200 low-quality road image sequences, each which consists of 31 consecutive frames captured by real dash cams. As a result, the RLPR dataset better reflects the image distortions found in real-world road scenarios compared to previous datasets, which consisted of high-resolution images and their artificially degraded versions. The RLPR dataset also includes pseudo ground-truth (GT) license plate images and text labels for each low-quality image sequence. The pseudo-GT image was created by extracting the license plate region from a high-quality frame within the same video clip as the low-quality image sequence, and then manually aligning it with the center frame among the 31 low-quality frames.

The main contributions of our work are summarized as follows:
\begin{itemize}
\item We report the inefficacy of conventional image restoration methods for license plate image restoration and address this limitation by proposing the MF-LPR² framework, which enhances legibility while preserving evidential content by leveraging multiple frames.
\item Optical flow filtering and refinement algorithms that precisely align multiple low-quality image frames by leveraging spatio-temporal consistency.
\item The realistic LPR dataset for evaluating multi-frame image restoration models under realistic conditions.
\item A novel metric, the Top-$k$ Percentile average of Distance to the Nearest Frame (PDNF-$k$) that quantifies the severity of spurious artifacts in the restored image.
\item Significant improvements in image quality and recognition accuracy compared to 11 baseline methods.
\end{itemize}

\section{Related Work}\label{Related Work}
\subsection{Image Restoration}
License plate image restoration aims to enhance low-quality license plate images to a level where accurate recognition is possible. Previous work on license plate image restoration primarily addresses the motion blur problem. In contrast, in the field of general image restoration, various methods, such as SISR, MFSR, and VSR, have been developed to enhance the resolution and visual quality of poor quality images. The majority of these methods are based on conditional generative models and restore missing information in input images using knowledge obtained from training data. In this section, we introduce restoration methods applicable to license plate images and describe their limitations in license plate image restoration. \\

\noindent{\textbf{Single Image Super-Resolution}}

In recent years, the field of SISR has seen significant progress, evolving from CNN-based architectures to Transformer-based approaches. SRCNN \cite{SRCNN} was one of the first deep-learning-based SR methods, utilizing a shallow CNN for high-resolution reconstruction. EDSR \cite{EDSR} improved upon this by removing batch normalization and increasing network depth. RCAN \cite{RCAN} introduced a Residual in Residual (RIR) structure and a channel attention mechanism, enabling deeper networks to focus on high-frequency information. More recent models, such as DCLS \cite{DCLS}, leverage deformable convolutions to adaptively enhance different regions of the image.

SRGAN \cite{SRGAN} improved perceptual quality by incorporating adversarial and perceptual loss functions. Subsequent models, including ESRGAN \cite{ESRGAN}, Real-ESRGAN \cite{Real-ESRGAN}, and BSRGAN \cite{BSRGAN}, further refined these methods by improving training stability and robustness to noise. Meanwhile, Transformer-based models have gained prominence: IPT \cite{IPT} applied self-attention mechanisms to SR tasks, SwinIR \cite{SwinIR} introduced Swin Transformers for better feature extraction, and Restormer \cite{Restormer} used channel-wise attention to boost efficiency. More recent methods, such as HAT \cite{HAT} and DRCT \cite{DRCT}, refine hierarchical attention strategies for high-fidelity image reconstruction.

Despite these notable advancements, existing SISR methods struggle to handle severe degradations and real-world noise characteristic of dash cam footage. Additionally, they are often trained on artificially degraded datasets, limiting their ability to generalize. To overcome these issues, our approach employs multi-frame information to surpass the inherent constraints of single-frame methods and validated on more realistic dataset. 

In addition to general super-resolution techniques, several studies have explored SR approaches specifically for text images. TPGSR \cite{TPGSR} introduced a text-prior-guided approach to enhance character restoration. STT \cite{STT} proposed a Transformer-based encoder for extracting text-specific features, while TATT \cite{TATT} and DPMN \cite{DPMN} introduced further refinements in text-specific feature extraction and reconstruction. However, these methods primarily address generic scene text and do not fully account for the unique challenges of license plates, such as severe motion blur, reflection light, and poor resolution. Our method specifically addresses license plate restoration by incorporating multi-frame alignment to handle severe degradation that is difficult to restore from a single image. \\

\noindent{\textbf{Multi-frame Super Resolution}}

MFSR models enhance image quality by aligning and aggregating information across consecutive frames. In recent models, optical flow estimators are commonly employed for frame alignment. Early optical flow-based techniques, such as the Lucas-Kanade algorithm \cite{Lucas-Kanade}, relied on handcrafted feature extractors. With the advent of deep-learning, CNN-based models like FlowNet \cite{FlowNet} and SpyNet \cite{SpyNet} demonstrated improved accuracy in flow estimation. More recently, Transformer-based approaches such as FlowFormer \cite{FlowFormer} extended SpyNet by integrating self-attention mechanisms, and FlowFormer++ \cite{FlowFormer++} further refined this architecture to boost flow estimation performance. While these approaches generally excel in multi-frame alignment, they often struggle under extreme low-quality conditions, leading to alignment errors and degraded final image quality. Building on optical flow-based alignment, our method introduces a novel spatio-temporal filtering and refinement mechanism for multi-frame aggregation, ensuring more accurate alignment and superior restoration results. \\

\noindent{\textbf{Video Super Resolution}}

VSR aims to reconstruct high-quality frames by leveraging information from multiple consecutive frames. VSR is distinguished from MFSR in that it restores multiple frames, whereas MFSR focuses on restoring a single 
 image. EDVR \cite{EDVR} introduced deformable convolutions to enhance feature aggregation, while BasicVSR++ \cite{BasicVSR++} advanced recurrent networks through improved temporal feature propagation. Later approaches, such as RVRT \cite{RVRT}, employed Transformer-based recurrent learning to refine temporal dependencies, and IART \cite{IART} further enhanced temporal consistency by reducing misalignment errors.
Although these methods exploit multi-frame information, they often rely on synthetic datasets where degradation is artificially induced, limiting their performance in real-world scenarios. Our approach addresses this challenge by integrating a spatio-temporal filtering and refinement mechanism that mitigates alignment errors and boosts robustness against real-world noise, particularly in dash cam footage.  \\

\subsection{License Plate Image Restoration and Recognition}
In this section, we introduce previous work on license plate recognition and image restoration techniques specially designed for license plate recognition. \\

\noindent{\textbf{License Plate Recognition}}

Lic ense plate recognition (LPR) involves both detection and character recognition. Numerous detection approaches have been proposed, ranging from general object detection frameworks such as Faster R-CNN \cite{FasterRCNN}, SSD \cite{SSD}, and YOLO \cite{YOLO}, to salient detection models like APNet Salient \cite{APNet}, WaveNet Salient \cite{WaveNet}, LSNet \cite{LSNet} which focus on identifying the most critical objects. Among these, LPR research has progressed in tandem with the development of general object detection methods. For instance, WPODNet \cite{WPOD}, built upon YOLO, was proposed for LPR but often struggles to deliver a reliable performance under real-world conditions involving low-quality license plate images. \\

\noindent{\textbf{License Plate Image Restoration}}

Previous work on license plate image restoration have mainly tackled motion deblurring \cite{CNN_Deblur, svoboda2016cnn, DeblurGAN, nguyen2018joint}.
Early studies on motion deblurring primarily estimated blur kernels using convolutional neural network (CNN)-based methods \cite{svoboda2016cnn}. Sun et al. \cite{CNN_Deblur} introduced a deep-learning framework explicitly designed to model non-uniform motion blur distributions at the patch level, setting a foundation for CNN-based motion deblurring techniques. Later, generative models such as  Kupyn et al. \cite{DeblurGAN} and Nguyen et al. \cite{nguyen2018joint} employed generative adversarial networks (GAN) to remove motion blur while preserving realistic textures. However, these models often struggle with recovering fine details in high-frequency areas such as small texts on license plates, leading to excessive smoothing and loss of crucial information.

Later, Zou et al. \cite{SemanticVLPSR} restored license plate images of extremely low-resolution by leveraging textual priors, similar to STISR methods. Nascimento et al. \cite{LPR-RSR-EXT} proposed a Transformer-based SISR method for low-resolution license plates, utilizing attention modules and an OCR-guided perceptual loss to enhance recognition accuracy. A more recent approach, AFA-Net \cite{AFANet}, integrated a super-resolution network and a deblurring network to achieve better restoration results. Nevertheless, these single-frame approaches do not take advantage of temporal information, which is critical for addressing severe degradation in real driving environments. Moreover, these methods were focused on artificially degraded datasets, limiting their ability to generalize to real-world noise. Our method extends these approaches by utilizing multi-frame information to achieve superior restoration performance.

The study most closely related to our work among previous research is `Eyes on the Target' \cite{EyesOnTarget}. However, it was developed long ago and relies on a traditional flow estimation algorithm to align frames, which limits its effectiveness in complex real-world scenarios. In contrast, our method exhibits significantly higher performance by incorporating a state-of-the-art neural flow estimator and novel flow filtering refinement methods. Table \ref{table:quantitavle_LPR_Quality} presents a comparative analysis demonstrating that MF-LPR² outperforms Eyes on the Target by large margins in terms of both image quality and recognition rate.

\begin{figure*}[t!]
\centering
\includegraphics[width=6.5in]{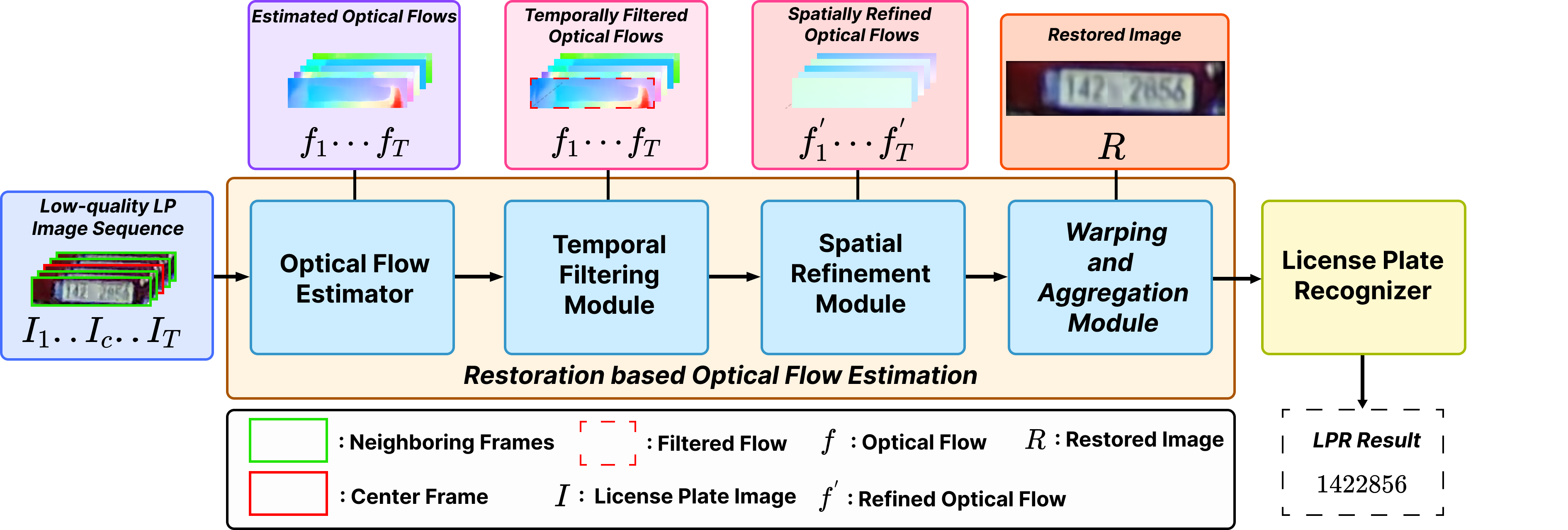}
\caption{The overall structure of the MF-LPR² framework.}
\label{over_all_framework}
\end{figure*}

\section{Method}\label{Method}
\subsection{Overview}
The proposed MF-LPR² restores and recognizes a high-quality license plate image from a sequence of low-quality image frames captured by consumer-grade cameras, such as dash cams. MF-LPR² consists of a restoration module, and a license plate recognizer, as illustrated in Fig. \ref{over_all_framework}.
Firstly, the restoration module restores the resolution and quality of the temporal center frame by combining the complementary information from its neighboring frames. Then, the license plate recognizer predicts the text from the restored image.

\subsection{Restoration Module}
The restoration module is a core component for recognizing low-quality license plate images. It enhances the temporal center frame, $I_c$, by aligning the neighboring frames ($I_t$'s with $t \neq c$) to $I_c$ and then aggregating the aligned frames to produce a high-quality image. For alignment, MF-LPR² estimates optical flow between the center frame and each neighboring frame, then warps the neighboring frames to match the center frame using the estimated optical flow. To precisely estimated optical flow, we combine a state-of-the-art optical flow estimator and novel filtering and refinement algorithms. The filtering algorithm rejects optical flows with severe errors based on temporal consistency, while the refinement algorithm detects and corrects local errors in the estimated optical flow based on spatial consistency.

\subsubsection{Optical Flow Estimator}

\begin{figure}[t!]
\centering
\includegraphics[width=4in]{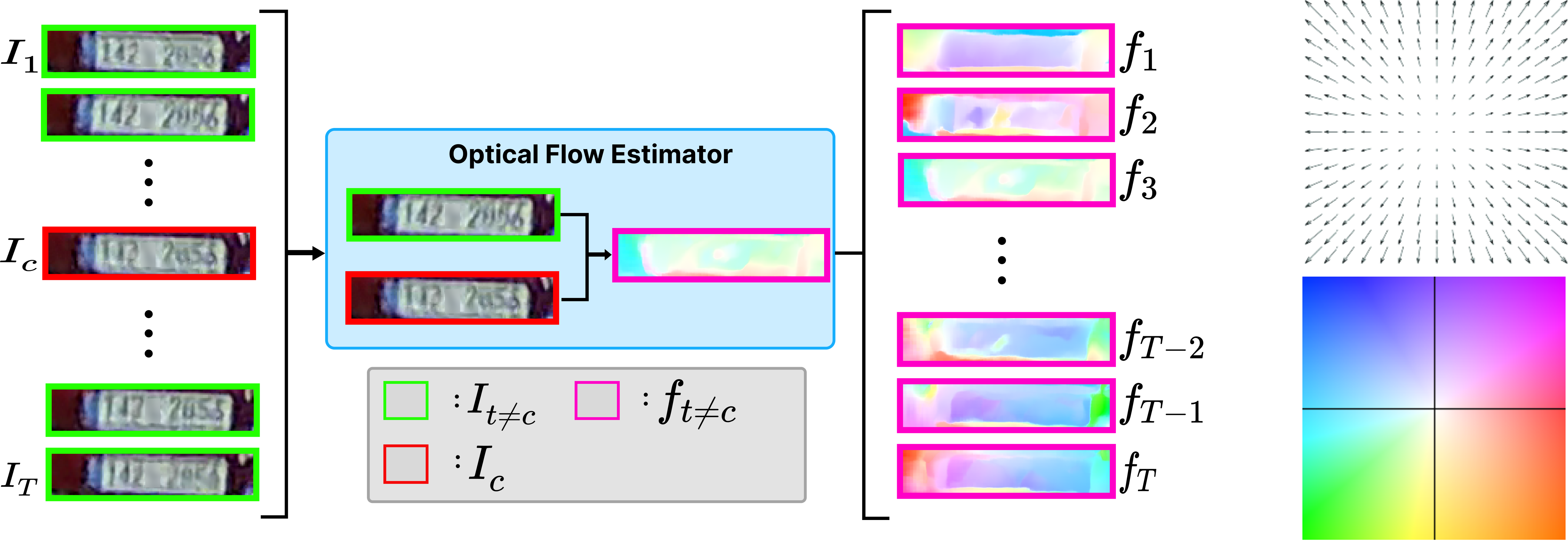}
\caption{Optical Flow Estimation: Given a sequence of license plate images, MF-LPR² estimates the optical flow from the center frame to each neighboring frame.
The direction represented by the colors of the optical flow corresponds to the color coding chart on the right.
}
\label{flow_error_viz1}
\end{figure}

In a dash cam video clip, the location and orientation of the license plate change over time. An optical flow represents the displacement of each pixel between a pair of frames. The estimator takes the center frame $I_c$ and a neighboring frame $I_t$ as input, and estimates optical flow $f_t$ from $I_c$ to each $I_t$, as Eq. \ref{eq:optical_flow}, where $H_c \times W_c$ is the size of $I_c$, $u_t \in \mathbb{R}^{H_c \times W_c}$ and $v_t \in \mathbb{R}^{H_c \times W_c}$ are the vertical and horizontal displacements from $I_c$ to $I_t$, respectively, i.e., $I_c(i, j) \approx I_t(i + u_t(i,j), j + v_t(i, j))$.
\begin{equation}
    f_t = (u_t, v_t) = \mathrm{OpticalFlow}(I_t,I_c)
\label{eq:optical_flow}
\end{equation}

\begin{figure*}[t!]
\centering
\includegraphics[width=6.5in]{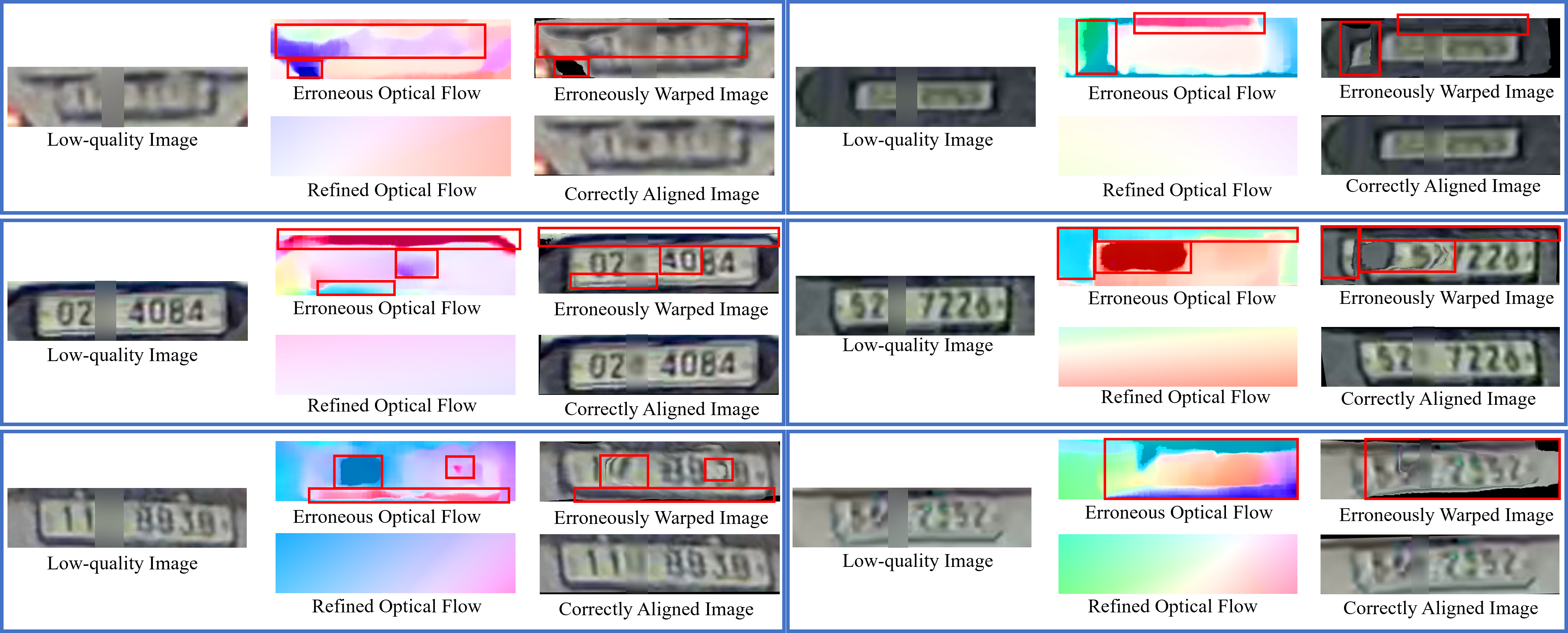}
\caption{ Comparison of erroneous and refined optical flows and their corresponding alignment results. Each blue box displays a low-quality image (left), the estimated and refined optical flows (center), and the alignment results based on the optical flows (right). The red boxes highlight errors. Even a state-of-the-art estimator, FlowFormer++ \cite{FlowFormer++}, produces substantial errors, which severely degrade the alignment results. However, the proposed method successfully refines the optical flows, significantly reducing the alignment errors as a result.
}
\label{fig:Spatial_diff_image}
\end{figure*}

Recently, Huang et al. \cite{FlowFormer} proposed a Transformer-based optical flow estimator, FlowFormer, which tokenizes a 4D cost volume from an input image pair, encodes the cost volume into a cost memory, and estimates optical flow by decoding the cost memory using a recurrent Transformer decoder. More recently, Shi et al. \cite{FlowFormer++} enhanced FlowFormer by pretraining the cost-volume encoder with a masked cost volume autoencoder, resulting in FlowFormer++. In this study, we apply the FlowFormer++ to estimate optical flows between the center and neighbor frames. Although FlowFormer++ is a state-of-the-art optical flow estimator that demonstrates high-performance on high-quality images, it often produces erroneous results on low-quality license plate images, as shown in Fig. \ref{fig:Spatial_diff_image}.

\subsubsection{Temporal Filtering Module}
The temporal filtering module rejects the optical flows that are poorly estimated overall based on temporal consistency between adjacent optical flows.
Since the interval between frames is short, the differences between adjacent optical flows are moderate. Therefore, we reject an optical flow $f_t$ if it differs significantly from the adjacent optical flows, i.e., $\mathrm{Average}(|f_{t+1} - f_t|) > \theta_{temp}$ and $\mathrm{Average}(|f_t - f_{t-1}|) > \theta_{temp}$, where $\theta_{temp}$ is a threshold value. In this study, we set $\theta_{temp} = 10$ according to the experimental result presented in Fig. \ref{fig:temporal_threshold}.
For the first and the last frames ($t=1$ and $t=T$), we compare $f_t$ with only one adjacent optical flow. When an optical flow $f_t$ is rejected, we discard the corresponding frame $I_t$. It is noteworthy that the temporal filtering module only rejects completely misestimated optical flows, and does not reject those that are partially misestimated, as rejections are based on the average difference.

We have attempted to refine misestimated optical flows rather than rejecting them. For example, we replaced misestimated optical flows $f_t$'s by the interpolation of the previous and next optical flows, $(f_{t-1} + f_{t+1}) / 2$. However, such refinement was ineffective in improving output quality because the frames with overall misestimated optical flows were usually too poor in quality to be used for restoring the center frame.

\subsubsection{Spatial Refinement Module}
Although the temporal filtering module rejects optical flows that are highly inconsistent with adjacent frames, the remaining optical flows still contain errors. Instead of aligning neighboring frames directly using the estimated optical flows, we refine them based on spatial consistency. Since a license plate is a rigid object with a planar shape, optical flows between license plate images are spatially smooth, this means that the values in $u_t$ and $v_t$ change linearly with the coordinates. Leveraging this spatial consistency, we refine the estimated optical flows to be planar through linear approximation, and use the resulting planar optical flow to align the frames.

To this end, we fit the estimated optical flow to a planar model. An important requirement for the fitting algorithm is that it should be robust to local errors in the estimated optical flow. In order to prevent local errors from affecting the refinement process, we fit the estimated optical flow to a plane based on the horizontal and vertical median values, which are tolerant to outlier values. First, we find the horizontal median values for each row and the vertical median values for each column, as Eq. \ref{eq:median}. As the spatial refinement is performed for each frame, we omit the time index $t$ for simplicity in this section.
\begin{equation}
    \label{eq:median}
    \begin{aligned}
        Med_{hor}(f)=(Med_{hor}(u), Med_{hor}(v))   \\
        Med_{ver}(f)=(Med_{ver}(u), Med_{ver}(v))
    \end{aligned}
\end{equation}
$Med_{hor}(u) \in \mathbb{R}^{H' \times 1}$ and $Med_{hor}(v) \in \mathbb{R}^{H' \times 1}$ are the horizontal median values of $u$ and $v$ for each row, while $Med_{ver}(u) \in \mathbb{R}^{1 \times W'}$ and $Med_{ver}(v) \in \mathbb{R}^{1 \times W'}$ are the vertical median values of $u$ and $v$ for each column. To further reduce the influence of outliers, we exclude the top and bottom 15\% of values and then find the median of the remaining values.

With the horizontal and vertical median values, we fit $Med_{hor}(u)$, $Med_{ver}(u)$, $Med_{hor}(v)$, and $Med_{ver}(v)$ to lines as Eq. \ref{eq:linear_approximation}, where $\alpha_*$ and $\beta_*$ are the coefficients for horizontal median values while $\gamma_*$ and $\delta_*$ are those for vertical median values. The coefficients are estimated from the median values by linear approximation.
\begin{equation}
    \label{eq:linear_approximation}
    \begin{aligned}
    R_{u}(i,j) &= (R_{hor}(u), R_{ver}(u)) = (\alpha_u i + \beta_u, \gamma_u j + \delta_u) \\
    R_{v}(i,j) &= (R_{hor}(v), R_{ver}(v)) = (\alpha_v i + \beta_v, \gamma_v j + \delta_v)
    \end{aligned}
\end{equation}

Then, we compute the refined optical flow $f'=(u',v')$ using Eq. \ref{eq:linear_approximation}, as Eq. \ref{eq:planar_optical_flow}.
\begin{equation}
    \label{eq:planar_optical_flow}
    \begin{aligned}
    f'(i,j) &= (u'(i,j), v'(i,j)) \\
    u'(i,j) &= \alpha_u (i - H/2) + \gamma_u j + \delta_u \\
    v'(i,j) &= \alpha_v (i - H/2) + \gamma_v j + \delta_v 
    % u'(i,j) &= \alpha_u i + \beta_u + \gamma_u (j - W/2) \\
    % v'(i,j) &= \alpha_v i + \beta_v + \gamma_v (j - W/2)
    \end{aligned}
\end{equation}

Fig. \ref{fig:Spatial_diff_image} presents examples of estimated and refined optical flows along with the corresponding alignment results. Each blue box displays a low-quality image (left), estimated and refined optical flows (center), and the alignment results (right). While the estimated optical flows contain substantial errors, producing erroneous alignment results, the proposed method demonstrates significantly improved results.

However, the refined optical flow $f'$ does not perfectly approximate the real optical flow because it is based on the assumption that the optical flow between license plate images is planar. Moreover, it estimates the coefficients from the median values of the estimated optical flow, which may contain error. Therefore, the difference between $f$ and the planar model $f'$, $\mathrm{Diff}(f,f')$, reflects not only estimation error in $f$ but also approximation error in $f'$, making the substitution of $f_t$ with $f'_t$ for all $t$ sub-optimal.

Therefore, we detect erroneous optical flows by comparing the maximum and median of $\mathrm{Diff}(f_t, f'_t)$ as Eq. \ref{eq:condition_to_replace}, where the maximum value represents the error in high-error areas, while the median value represents the error in other areas. The former reflects estimation error, whereas the latter reflects approximation error. We set $\theta_{spatial}=20$ according to the experimental result presented in Fig \ref{fig:spatial_threshold}.
\begin{equation}
    \label{eq:condition_to_replace}
    \mathrm{max}_{i,j}(\mathrm{Diff}(f(i,j), f'(i,j))) - \mathrm{median}_{i,j}(\mathrm{Diff}(f(i,j), f'(i,j))) > \theta_{spatial}
\end{equation}

\subsubsection{Warping and Aggregation Module}
With the refined optical flows, we align and aggregate the neighboring frames to produce a high-quality image. For this, we apply the Geometric k-nearest neighbors Super-Resolution (GSR$_4$) algorithm \cite{GSR}. First, it places the pixels from all frames $I_t$ onto the coordinate plane of the output image, which is the same as that of the center frame in MF-LPR² except that it allows fractional coordinates. It transforms the coordinate of each pixel in $I_t$ to the corresponding output coordinate using the optical flow. Then, for each integer coordinate $(i, j)$ on the output coordinate plane, the it finds the nearest pixels in each frame. Finally, it fills each output pixel $\hat{Y}(i,j)$ with the average of the nearest neighbor pixels for each $(i, j)$.

It is noteworthy that, since GSR$_4$ determines pixel values based on the average of aligned neighboring frames, it effectively suppresses noise and outlier pixels without introducing structural patterns that deviate significantly from those of the input images.
As a result, as shown in Fig. \ref{fig:visual_comparison}, MF-LPR² does not introduce severe artifacts or distortions, thereby avoiding arbitrary alterations to the evidential contents of the input images. In license plate recognition, this represents a significant advantage compared to generative models that actively utilize pretrained priors to restore images.

\subsection{License Plate Recognizer}
To segment and recognize the characters in the restored license plate image, we utilize an off-the-shelf scene text recognizer, MGP-STR \cite{MGP-STR}, which is both efficient and effective in handling low-quality images. MGP-STR demonstrates excellent performance in managing diverse text appearances, varying orientations, and complex backgrounds, making it particularly well-suited for recognizing characters in license plate images captured in real-world scenarios.

\subsection{The Realistic License Plate Restoration and Recognition Dataset}

Since MF-LPR² takes a sequence of road image frames as input, its evaluation requires a dataset of low-quality road image sequences that resemble real-world driving conditions. Additionally, for quantitative evaluation, a high-quality pseudo ground truth (GT) image and text label for each image sequence are essential. However, existing license plate datasets do not satisfy these specific requirements. \newline

\noindent{\textbf{Limitations of Existing Datasets}}

Several existing license plate datasets, such as OpenALPR-EU \cite{OpenALPR-EU}, SSIG-SegPlate \cite{SSIG-SegPlate}, UFPR-ALPR \cite{UFPR-ALPR}, PKU-SR \cite{PKU-SR/Rodosol-SR}, and RodoSol-SR \cite{PKU-SR/Rodosol-SR}, have been introduced primarily for Automatic License Plate Recognition (ALPR). While these datasets feature images from diverse environments, they are not well-suited for evaluating multi-frame restoration algorithms for low-quality license plate images. OpenALPR-EU focuses on single-frame images of parked vehicles, which fail to represent real-world driving conditions. SSIG-SegPlate and UFPR-ALPR include multi-frame sequences but generally comprise high-quality images, not reflecting the challenges of real-world low-quality data. Datasets like PKU-SR and RodoSol-SR rely on artificial degradation methods, such as Gaussian noise and bicubic downsampling, to generate low-quality frames.

\begin{figure*}[t!]
\centering
\includegraphics[width=6.5in]{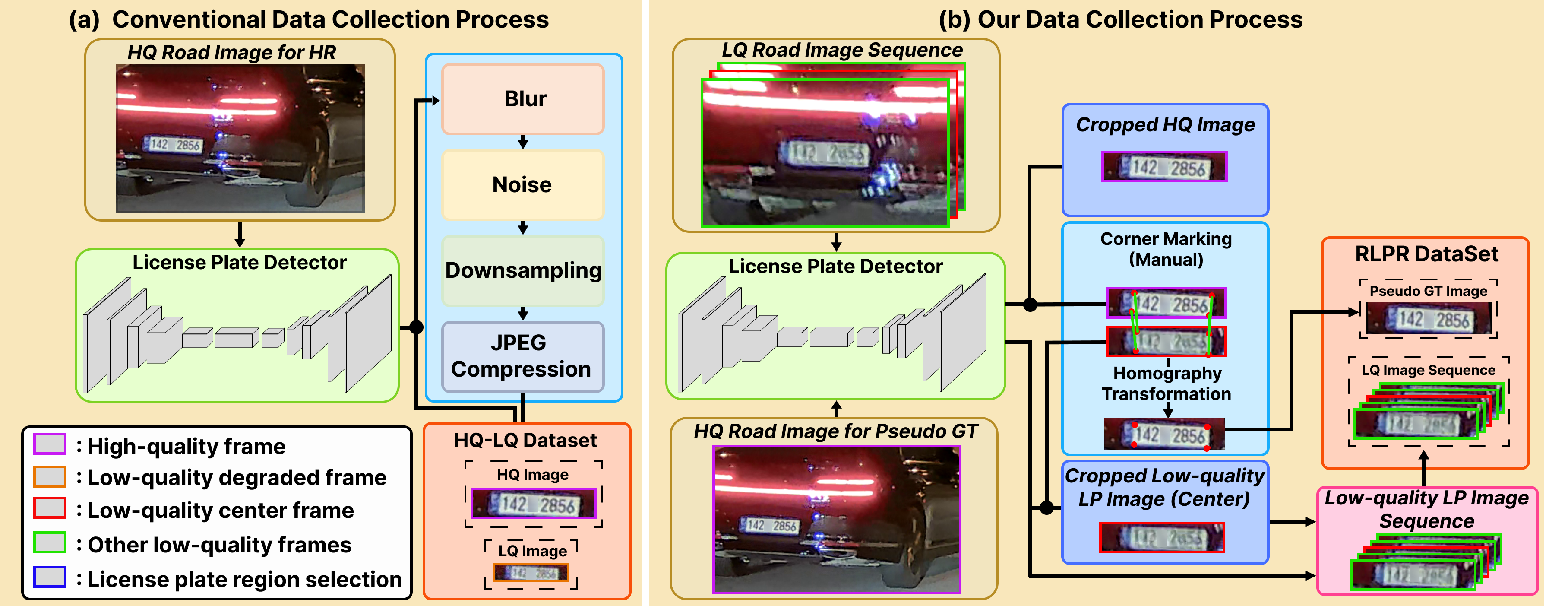}
\caption{The process of collecting high- and low-quality image pairs: the conventional process based on artificial degradation (Left) and the process used to collect the RLPR dataset (Right). The RLPR dataset consists of real low-quality images that reflect the complexities of real-world scenarios. }
\label{RLPR_Framework}
\end{figure*}

Fig. \ref{RLPR_Framework} (a) illustrates the conventional process of generating high- and low-quality image pairs through artificial degradation. This process applies multiple degradation steps, including noise addition, downsampling, and lossy compression. However, artificially degraded images often fail to mimic the diverse and complex degradations found in real-world conditions. Furthermore, the consistent level of degradation across all frames reduces inter-frame complementarity, making these datasets less effective for evaluating multi-frame restoration algorithms. \newline

\noindent{\textbf{Dataset Construction Process}}

The building process of the RLPR dataset (as illustrated in Fig. \ref{RLPR_Framework} (b)) ensures that the dataset closely mirrors real-world driving conditions while maintaining high-quality annotations. From 1,052 dash cam video clips, we extracted sequences with visible license plates exhibiting sufficiently low quality. We also extracted a high-quality frame for the pseudo-GT image from the same video clip. Only 200 sequences were retained after meticulous manual screening to ensure they included both low-quality frames and high-quality frames. For each sequence, the following steps were performed:

First, a state-of-the-art DeepLabV3 \cite{Deeplabv3} model with a ResNet-101 backbone, fine-tuned on 130 manually annotated road images, was used to detect license plate regions. These regions were manually refined and cropped with slightly expanded bounding boxes to preserve important details, resulting in slight positional and size variations across frames that reflect real-world conditions.

Pseudo-GT images were then selected as the highest-quality frames from the same video clips but were often temporally distant from the low-quality sequences. To ensure precise alignment, the corners of the license plate regions were manually marked on both the pseudo-GT and the center frame of the sequence, followed by a homography transformation. The aligned GT images retained only the license plate region, with background areas removed, ensuring accurate comparability with the low-quality frames.

This rigorous construction process ensures that RLPR captures the complexities of real-world conditions, overcoming the limitations of artificially degraded datasets. Unlike other datasets that rely on artificial degradation, RLPR provides sequences with diverse, dynamic degradations and complementary frame information, making it uniquely suited for evaluating multi-frame restoration frameworks. %\newline

The RLPR dataset encompasses a diverse range of conditions, including lighting (daytime and nighttime), lane positions (ego and non-ego lanes), and license plate types (painted and reflective), as shown in Table~\ref{table:dataset_detail}. This diversity ensures that RLPR serves as a robust benchmark for realistic performance evaluation. Furthermore, by pairing each sequence with carefully aligned high-quality pseudo-GT images, RLPR enables researchers to perform precise quantitative analysis of restoration and recognition tasks. 
Consequently, despite the limited number of samples, which makes the RLPR dataset not suitable for model training, it possesses high value as an evaluation dataset.

\begin{table}[ht]
\caption{Composition of RLPR dataset. Each sample is categorized by lighting condition (daytime/nighttime), lane position (ego/non-ego), and plate type (painted/reflective film).}
\centering % used for centering table
\normalsize
\begin{tabular}{c c c} % centered columns 
\toprule
Category & Number of Samples & Number of Characters\\
\hline
\midrule
\textbf{RLPR Dataset} & 200 & 1261\\
\hline
% \multicolumn{3}{l}{\textbf{Category 1: Daytime / Nighttime}} \\  % Use \multicolumn to span across all columns
Daytime & 119 & 715 \\
Nighttime & 81 & 510 \\
\hline
% \multicolumn{3}{l}{\textbf{Category 2: Ego Lane / Non-ego Lane}} \\  % Use \multicolumn to span across all columns
Ego Lane & 42 & 263 \\
Non-ego Lane & 158 & 998 \\
\hline
% \multicolumn{3}{l}{\textbf{Category 3: Painted / Reflective Film}} \\  % Use \multicolumn to span across all columns
Painted & 165 & 1016 \\
Reflective Film & 35 & 245 \\
\bottomrule
\end{tabular}
\label{table:dataset_detail} % is used to refer this table in the text
\end{table}

\subsection{Top-$k$ Percentile Average Distance to Nearest Frame (PDNF-$k$)}
\label{subsec:PDNF}
A critical and unique requirement in license plate image restoration is that the evidential contents of the original image must not be compromised. However, many restoration models, which actively utilize priors obtained from training data, frequently generate spurious artifacts, as shown in Fig. \ref{fig:visual_comparison}. The artifacts significantly degrade the essential information required to preserve character class features.

Nevertheless, since artifacts occupy only a small portion of the overall image, conventional quality assessment metrics that reflect the global difference with the GT image fail to adequately capture the impact of these artifacts.
Common metrics like PSNR, SSIM, and LPIPS integrate local differences via spatial averaging or Minkowski pooling. PSNR is derived from the mean squared error (MSE) between the GT and output images. SSIM measures structural similarity through Minkowski pooling across spatial and frequency domains. LPIPS evaluates perceptual similarity by averaging embedding differences across spatial dimensions and layers. While these metrics capture overall differences, they are not sensitive to localized spurious artifacts.

To complement such a limitation of conventional metrics, we propose a novel metric to quantify the degradation of evidential contents caused by spurious artifacts.
An effective metric that quantifies the severity of spurious artifacts must meet the following conditions:

\begin{enumerate}
    \item To measure the degradation of original information rather than restoration performance, it should be computed based on the difference from the input image, rather than the GT image.

    \item Since low-quality input images reflect pixel values with uncertainty, the metric should account for this by representing each pixel value as a distribution. This helps mitigate unnecessary penalties that correctly restored images may otherwise receive.

    \item It should focus on regions with significant discrepancies rather than simply integrating local differences across spatial dimensions.
\end{enumerate}

Regarding conditions 1 and 2, we estimate the distribution of each input pixel $(i, j)$ based on the corresponding pixel values from multiple frames. However, modeling the distribution of pixel values as a Gaussian distribution is inappropriate for two reasons. First, when the aligned frames are similar to each other, the variation is measured to be excessively low, leading to an underestimation of pixel uncertainty. Second, since $\mathrm{GSR}_4$ applied to MF-LPR² outputs the mean value, a Gaussian model would overly favor MF-LPR². Therefore, we adopt a non-parametric approach similar to the k-nearest neighbor classifier. Specifically, the error for each output pixel $(i, j)$ is defined as the distance to the nearest corresponding pixels in the aligned input frames.
\begin{align}
    \mathrm{Dist}_t(i, j) &= \left\lvert \hat{Y}(i,j) - I'_t(i,j) \right\rvert \\
    \mathrm{DNF}(i, j) &= \mathrm{min}_t \mathrm{Dist}_t(i, j),
\end{align}
where $I'_t$ is the $t$-th aligned input frame as $I'_t(i,j) = I_t(i + u_t(i,j), j + v_t(i,j))$.

For condition 3, we integrate pixel-wise errors DNF$(i, j)$ using the top-$k$ percentile average as
\begin{equation}
    \mathrm{PDNF\text{-}}k = \frac{1}{|\Omega_k|} \sum_{(i,j) \in \Omega_k} \mathrm{DNF}(i,j),
\end{equation}
where $\Omega_k$ is the set of pixels with the highest \(k\%\) DNF values.

The key characteristics of $\mathrm{PDNF\text{-}}k$ are as follows:
\begin{itemize}
    \item A large PDNF-$k$ value indicates that the output image contains severe spurious artifacts.

    \item A smaller PDNF-$k$ value suggests that the information in the original image is better preserved. However, an excessively small PDNF-$k$ value may imply a limited enhancement effect on output quality. For instance, if the model produces an output identical to the low-quality center frame, PDNF-$k$ is measured as zero.

    \item PDNF-$k$ is a metric designed to complement existing quality assessment metrics and is intended to be used alongside other metrics.

    \item As $k$ decreases, PDNF-$k$ focuses more on pixels with large errors, whereas as $k$ increases, it reflects overall error.
\end{itemize}

\section{Experiments}
\subsection{Experimental Settings and Evaluation Metrics}
Our framework consists of two main components: a restoration module and a license plate recognizer. For the optical flow estimator in the restoration module and for the license plate recognizer, we used the official implementations of FlowFormer++ \cite{FlowFormer++-opensource} and MGP-STR \cite{MGP-STR-opensource}, respectively. All experiments were conducted on a server equipped with an NVIDIA RTX 2080.

For evaluation, we first restored and recognized license plate areas in the RLPR dataset by MF-LPR². We then evaluated the image restoration performance by comparing the restored images with the pseudo-GT images. For quantitative evaluation, we assessed the quality of the restored images using metrics such as Peak Signal-to-Noise Ratio (PSNR), Structural Similarity Index (SSIM) \cite{SSIM}, and Learned Perceptual Image Patch Similarity (LPIPS) \cite{LPIPS}. These metrics were computed only for the pixels within the license plate region, excluding the background areas. Additionally, we measured the severity of spurious artifacts in PDNF-$5$ introduced in Section \ref{subsec:PDNF}.
Finally, we evaluated the recognition performance by measuring the character recognition accuracy using the text labels from the RLPR dataset.
For comparison analysis, we compared MF-LPR² with eight recently-developed image/video restoration models, (SwinIR, DCLS, HAT, DRCT, DPMN, TATT, RVRT, and IART) and three license plate recognition (LPR) models (WPOD-Net, AFA-Net, and Eyes on the Target). The performance of the baseline models was measured using their open-source implementations and pretrained parameters.

\subsection{Comparison with Image/Video Super-Resolution Models}

\begin{table}[ht]
\caption{Image quality evaluation results of the proposed model, MF-LPR², compared with other recently developed baseline models. $\uparrow$ indicates higher values are better, while $\downarrow$ indicates lower values are better. The best and second-best performances are highlighted in bold and underlined, respectively. MF-LPR² outperformed the baseline models in all metrics.}
\centering % used for centering table
\normalsize
\begin{tabular}{c c c c c c c} % centered columns 
\toprule
Method & Single/Multi Frame & PSNR$\uparrow$ & SSIM$\uparrow$ & LPIPS$\downarrow$ & PDNF-5$\downarrow$ & LPR Accuracy$\uparrow$\\
\hline
\midrule
SwinIR \cite{SwinIR} & Single Frame & 14.6156 & 0.2931 & 0.5495 & 108.12 & 3.49\%\\
DCLS \cite{DCLS} & Single Frame & 15.6955 & 0.3121 & 0.5908 & 27.60 & 16.18\%\\
HAT \cite{HAT} & Single Frame & 15.6339 & 0.3064 & 0.6021 & 35.50 & 15.07\%\\
DRCT \cite{DRCT} & Single Frame & 15.6031 & 0.3023 & 0.5942 & 37.94 & 15.07\%\\
TATT \cite{TATT} & Single Frame & 14.8347 & 0.3149 & 0.5393 & 69.80 & 9.36\%\\
DPMN \cite{DPMN} & Single Frame & 15.6304 & \underline{0.3383} & \underline{0.5036} & 40.11 & 14.99\%\\
\hline
RVRT \cite{RVRT} & Multi Frame & 15.7743 & 0.3155 & 0.5595 & \textbf{6.96} & 18.56\%\\
IART \cite{IART} & Multi Frame & \underline{15.7951} & 0.3149 & 0.5565 & \underline{9.58} & \underline{18.95\%}\\
\textbf{MF-LPR²} & Multi Frame & \textbf{16.3478} & \textbf{0.3486} & \textbf{0.4629} & 14.00 & \textbf{86.44\%}\\
\bottomrule
\end{tabular}
\label{table:quantitative_SR_Quality} % is used to refer this table in the text
\end{table}

We compared the performance of MF-LPR² against eight state-of-the-art image and video super-resolution models: SwinIR\cite{SwinIR}, DCLS\cite{DCLS}, HAT\cite{HAT}, DRCT\cite{DRCT}, DPMN\cite{DPMN}, TATT\cite{TATT}, RVRT\cite{RVRT}, and IART\cite{IART}. The evaluation was carried out using PSNR, SSIM, LPIPS, PDNF-$5$ and character recognition accuracy metrics. For character recognition accuracy, we integrated each restoration model with the same recognizer used for MF-LPR².

Table \ref{table:quantitative_SR_Quality} presents the evaluation results. The poor quality of input images caused the baseline models to produce unreliable results, often leading to spurious artifacts and low recognition accuracy. Fig. \ref{fig:visual_comparison} illustrates the restoration results of several samples.
MF-LPR² showed significantly higher performance in all metrics, achieving a PSNR of 16.3478, a SSIM of 0.3486, and a LPIPS of 0.4629. Notably, MF-LPR² demonstrated a character recognition accuracy of 86.44\% which is remarkably higher than the second best performance of 18.95\% achieved by IART. These results show the effectiveness of MF-LPR² in restoration and recognition of poor quality license plate images.

Regarding PDNF-$5$, MF-LPR² showed the third lowest value, following RVRT and IART. As shown in Fig. \ref{fig:visual_comparison}, these three models do not generate spurious artifacts significantly. However, RVRT and IART produced output images that overly resemble the input images. As a result, they performed worse than MF-LPR² in other metrics. SwinIR, TATT, and DPMN exhibited significantly high PDNF-$5$ values indicating that they suffer severely from artifacts, as shown in Fig. \ref{fig:visual_comparison}.

\begin{table}[ht]
\setlength{\tabcolsep}{4pt} % Adjust column spacing (default is 6pt)
\caption{Evaluation results of MF-LPR² compared with a license plate image restoration and recognition framework, WPOD-Net, AFA-Net, and Eyes on the Target. $\uparrow$ indicates higher values are better, while $\downarrow$ indicates lower values are better. MF-LPR² outperformed the baseline models in all metrics. As WPOD-Net model is a recognition model without a restoration module, its SSIM, PSNR, and LPIPS calculation are omitted. The high PDNF-$5$ value of Eyes on the Target is due to the official code, which generates rectified images that may be misaligned relative to the input image.}
\centering % used for centering table
\normalsize
\begin{tabular}{c c c c c c c} % centered columns 
\toprule
Method&Single/Multi Frame&PSNR$\uparrow$&SSIM$\uparrow$&LPIPS$\downarrow$ & PDNF-5$\downarrow$ & LPR Accuracy$\uparrow$\\  % inserts table
\hline
\midrule
WPOD-Net \cite{WPOD} & Single Frame & - & - & - & - & 13.08\%\\
AFA-Net \cite{AFANet} & Single Frame & 14.7506 & 0.3018 & 0.6110 & 87.85 & 14.04\%\\
\hline
Eyes on the Target \cite{EyesOnTarget} & Multi Frame & 14.9170 & 0.2680 & 0.5126 & 117.77 & 82.55\%\\
\textbf{MF-LPR²} & Multi Frame & \textbf{16.3478} & \textbf{0.3486} & \textbf{0.4629} & \textbf{14.00} & \textbf{86.44\%}\\
\bottomrule
\end{tabular}
\label{table:quantitavle_LPR_Quality} % is used to refer this table in the text
\end{table}

\subsection{Comparison with License Plate Image Restoration and Recognition Models}
We also compared MF-LPR² with three license plate image restoration and recognition models presented in previous studies, WPOD-Net\cite{WPOD}, AFA-Net\cite{AFANet}, and Eyes on the Target\cite{EyesOnTarget}, which were specifically designed for license plate images. Table \ref{table:quantitavle_LPR_Quality} presents the comparison results.
The single-frame LPR frameworks, WPOD-Net and AFA-Net, achieved an LPR accuracy of 13.08\%, showing no significant improvement compared to MGP-STR combined with the restoration models, shown in Table \ref{table:quantitative_SR_Quality}.

However, the multi-frame LPR model, Eyes on the Target, achieved significantly higher character recognition accuracy than the single-frame LPR frameworks. MF-LPR² exhibited even higher performance than Eyes on the Target across all metrics largely due to its accurate estimation of optical flows via the proposed filtering and refinement algorithms. MF-LPR² also outperformed all baseline LPR models in PSNR, SSIM, and LPIPS by large margins.

\begin{figure*}[t!]
\centering
\includegraphics[width=6.5in]{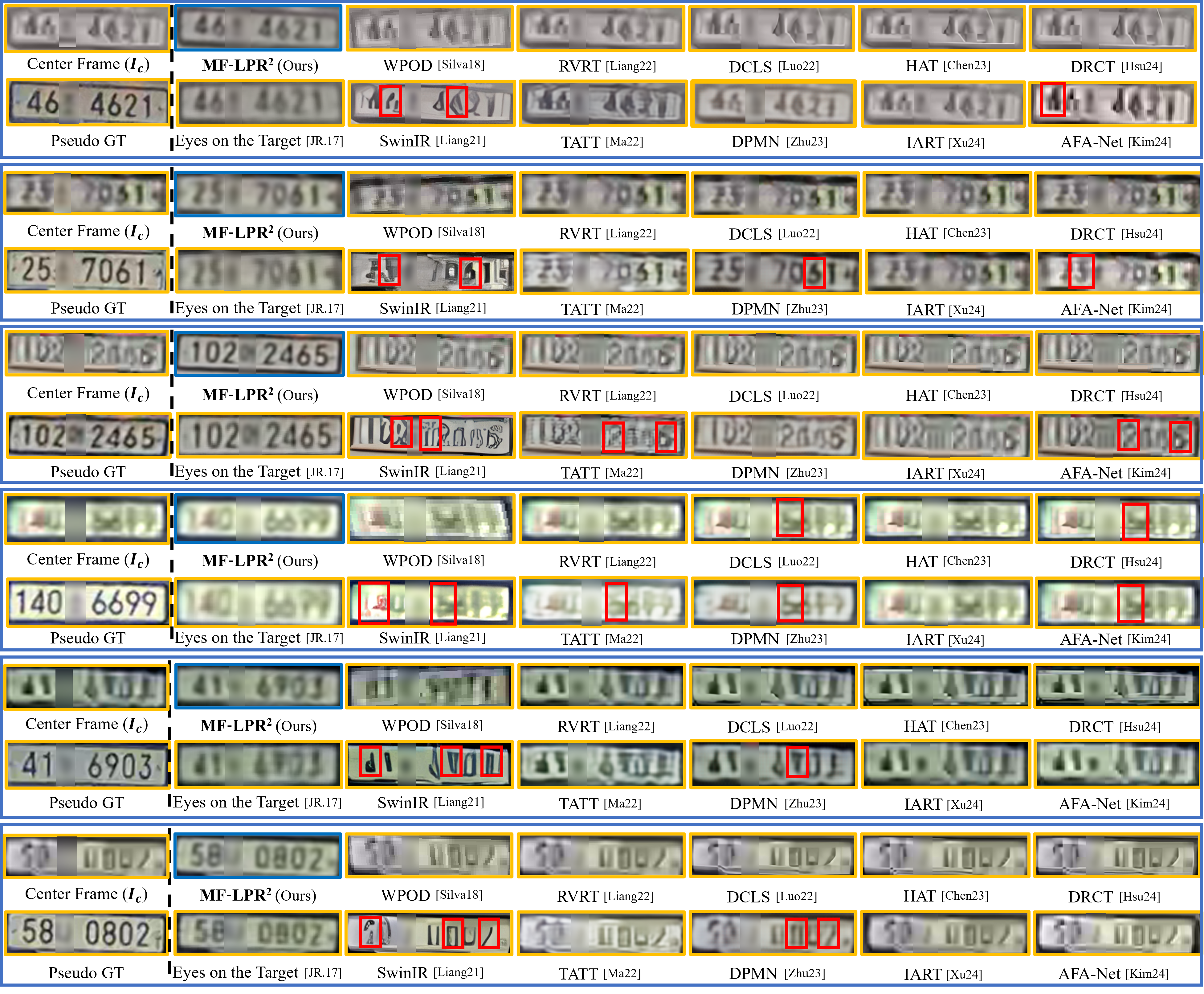}
\caption{The restoration results of MF-LPR² compared with baseline models. The red boxes highlight artifacts and distortions due to the poor quality of the input image. We blurred the third letter in each image to de-identify personal information.}
\label{fig:visual_comparison}
\end{figure*}

\begin{figure*}[t!]
\centering
\includegraphics[width=6in, height=2.5in]{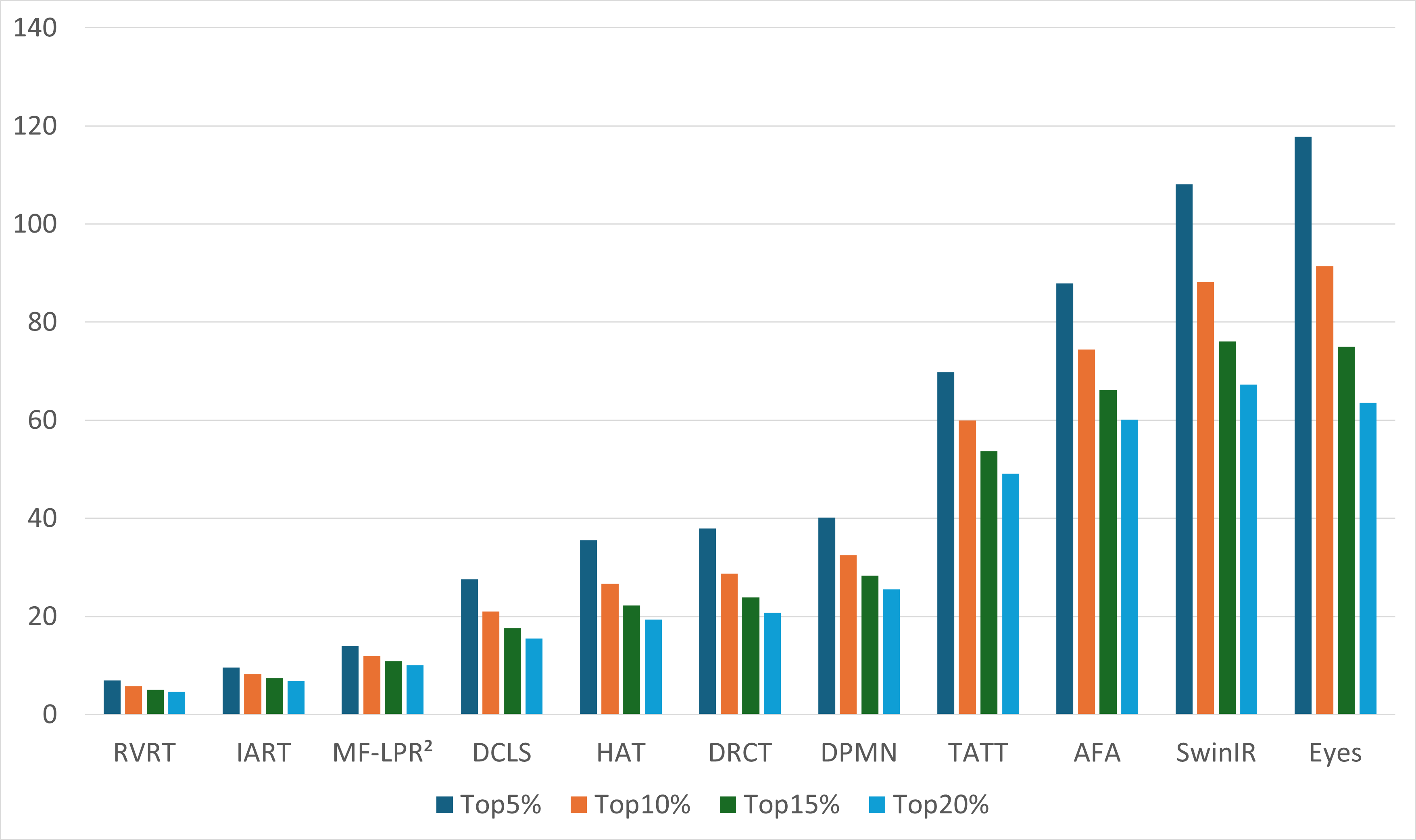}
\caption{The PDNF-$k$ of restoration models averaged over the RLPR dataset with $k = 5, 10, 15, 20$.}
\label{fig:pdnf-k_graph}
\end{figure*}

MF-LPR² also exhibited a lower PDNF-$5$ value than the other models. Eyes on the Target showed a high PDNF-$5$ value, while its results in Fig. \ref{fig:visual_comparison} contain little artifacts. The high PDNF-$5$ value is attributed to the official code, which generates rectified images that may be misaligned relative to the input image. This may also affect the PSNR value, whereas SSIM, LPIPS, and LPR accuracy are less sensitive to slight misalignment. For the reliability of the experiment, we did not modify the official code. 
The PDNF-$k$ values of the models for $k = 5, 10, 15, 20$ are compared in Fig. \ref{fig:pdnf-k_graph}.
The graph exhibits a similar trend for all $k$. However, the differences between models are most pronounced in PDNF-$5$, which focuses the most on severe artifacts.

\subsection{Qualitative Evaluation}

In addition to the quantitative evaluation using the pseudo-GT images, we qualitatively compared the restoration results of MF-LPR² with those of the eleven baseline models, SwinIR, DCLS, HAT, DRCT, DPMN, TATT, RVRT, IART,  WPOD-Net, AFA-Net, and Eyes on the Target. Fig. \ref{fig:visual_comparison} displays the restoration results of MF-LPR² and the baseline restoration methods. When applied to low-quality images, the super-resolution models did not effectively enhance their visual quality. Especially, they often produced spurious artifacts as highlighted in the red boxes. For example, the character `6' in the second input image was severely degraded, making it easily confused with `0', `5', or `3.' The eight baseline methods (SwinIR, DCLS, HAT, DRCT, DPMN, TATT, RVRT, and IART) did not reduce ambiguity but restored it similarly to character `5' or `3.' The output quality of Eyes on the Target was the best among the baseline models, but it produced more blurry outputs compared to MF-LPR². In contrast, the proposed model, MF-LPR², yielded significantly superior results by reducing ambiguity and enhancing visual readability.

\subsection{Model Complexity}

\begin{table}[ht]
\setlength{\tabcolsep}{4pt} % Adjust column spacing (default is 6pt)
\caption{Complexity Analysis between License Plate Image Restoration and Recognition Models. Inference Time is calculated per frame.}
\centering % used for centering table
\normalsize
\begin{tabular}{c c c c c} % centered columns 
\toprule
Method & Single/Multi Frame & Params (M) & FLOPs (G) & Inference Time (s)\\  % inserts table
\hline
\midrule
WPOD-Net \cite{WPOD} & Single Frame & 1.64 & 3.73 & 0.18\\
AFA-Net \cite{AFANet} & Single Frame & 139.64 & 481.62 & 0.55\\
\hline
Eyes on the Target \cite{EyesOnTarget} & Multi Frame & - & - & 3 \\
\textbf{MF-LPR²} & Multi Frame & 163.76 & 71.57 & 5.7 \\
\bottomrule
\end{tabular}
\label{table:complexity_LPR} % is used to refer this table in the text
\end{table}

We measured the complexity of the license plate restoration and recognition models. Table \ref{table:complexity_LPR} summarizes the comparison results for these models. The number of parameters and the amount of compute in FLOPs were measured using the PyTorch Profiler. WPOD-Net, a single frame-based recognition model, is the lightest and fastest in terms of computational cost among single frame models; however, its performance is significantly lower than other models. In contrast, AFA-Net incorporates both a super-resolution network and a deblurring network, resulting in the highest FLOPs among all models despite being single frame-based, yet its performance remains limited. For Eyes on the Target — a framework based on traditional algorithms — only the inference time was evaluated. Our proposed MF-LPR² model strikes a reasonable balance by achieving a relatively low FLOPs-to-parameter ratio. While its inference time is somewhat higher compared to other models, it demonstrates superior overall performance, making it an effective solution for license plate restoration and recognition tasks.

\begin{figure}[ht]
\centering
\includegraphics[width=4in]{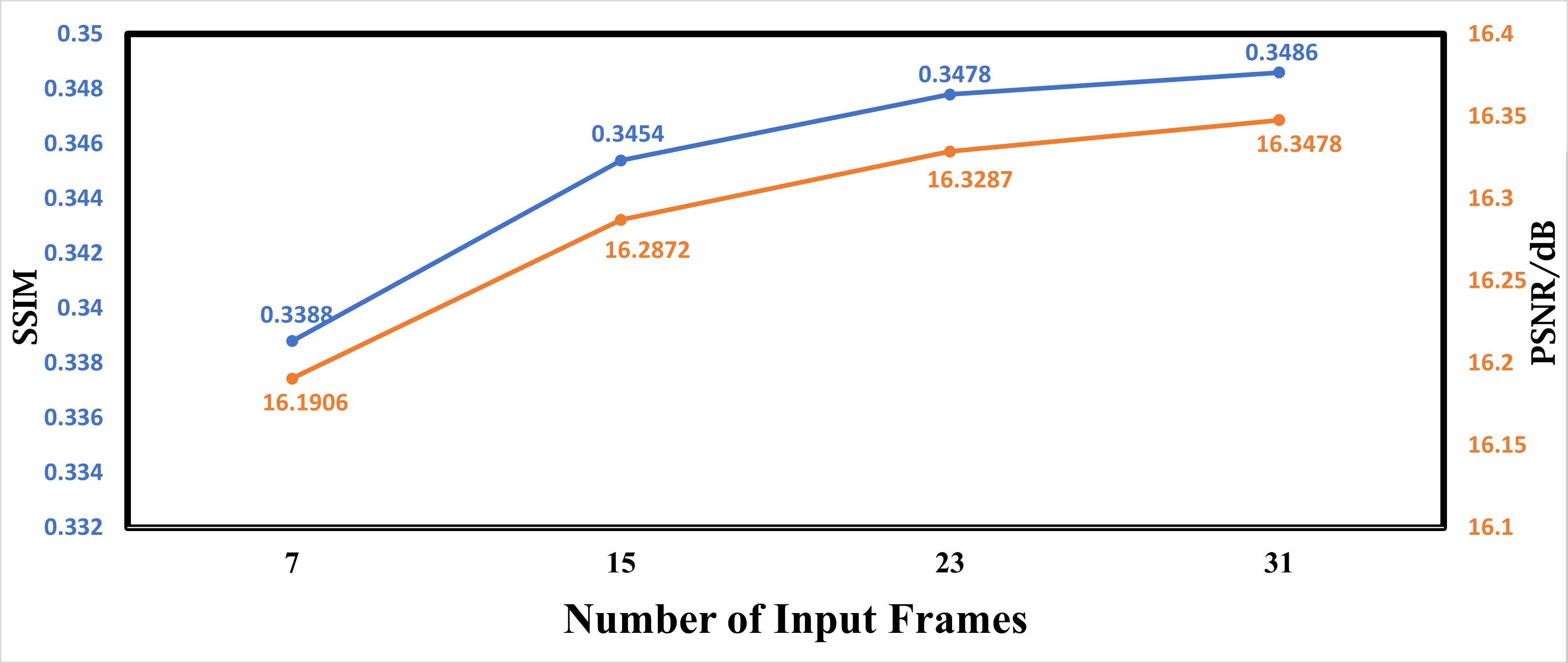}
\caption{The performance of MF-LPR² with varying numbers of input frames.}
\label{fig:ref_images}
\end{figure}

\subsection{Ablation Studies}
We performed ablation studies to analyze the impact of the choices for each component on performance.
We compared the performance of MF-LPR² by modifying various design options such as the number of frames in the input sequence, existence of modules, the threshold values for the temporal filtering and spatial refinement of the optical flow.

\subsubsection{The Number of Frames in Input Image Sequence}
We evaluated the restoration performance of MF-LPR² in SSIM and PSNR by changing the number of low-quality images in the input sequence. Fig. \ref{fig:ref_images} displays the average SSIM and PSNR values of the output images restored from  7, 15, 23, and 31 input frames. Both SSIM and PSNR values consistently increase with the number of input frames. These results suggests that MF-LPR² effectively utilizes the information of multiple input frames to enhance the quality of the center frame.

\begin{table}[ht]
\caption{Performance comparison of MF-LPR² with different module configurations. \textbf{F}, \textbf{T}, and \textbf{S} represent optical flow estimation, temporal filtering, and spatial refinement, respectively. Both temporal filtering and spatial refinement improve PSNR, SSIM, and recognition accuracy.}
\centering
\normalsize
\begin{tabular}{c c c c c c}
\toprule
F & T & S & PSNR$\uparrow$ & SSIM$\uparrow$ &  LPR Accuracy$\uparrow$  \\  
\midrule
\checkmark & & & 16.3114 & 0.3468 & 74.70\% \\
\checkmark & \checkmark & & 16.3240 & 0.3480 & 75.97\% \\
\checkmark & \checkmark & \checkmark & \textbf{16.3478} & \textbf{0.3486} & \textbf{86.44\%} \\ 
\bottomrule
\end{tabular}
\label{table:module_effectiveness}
\end{table}

\begin{figure}[ht]
\centering
\includegraphics[width=6in]{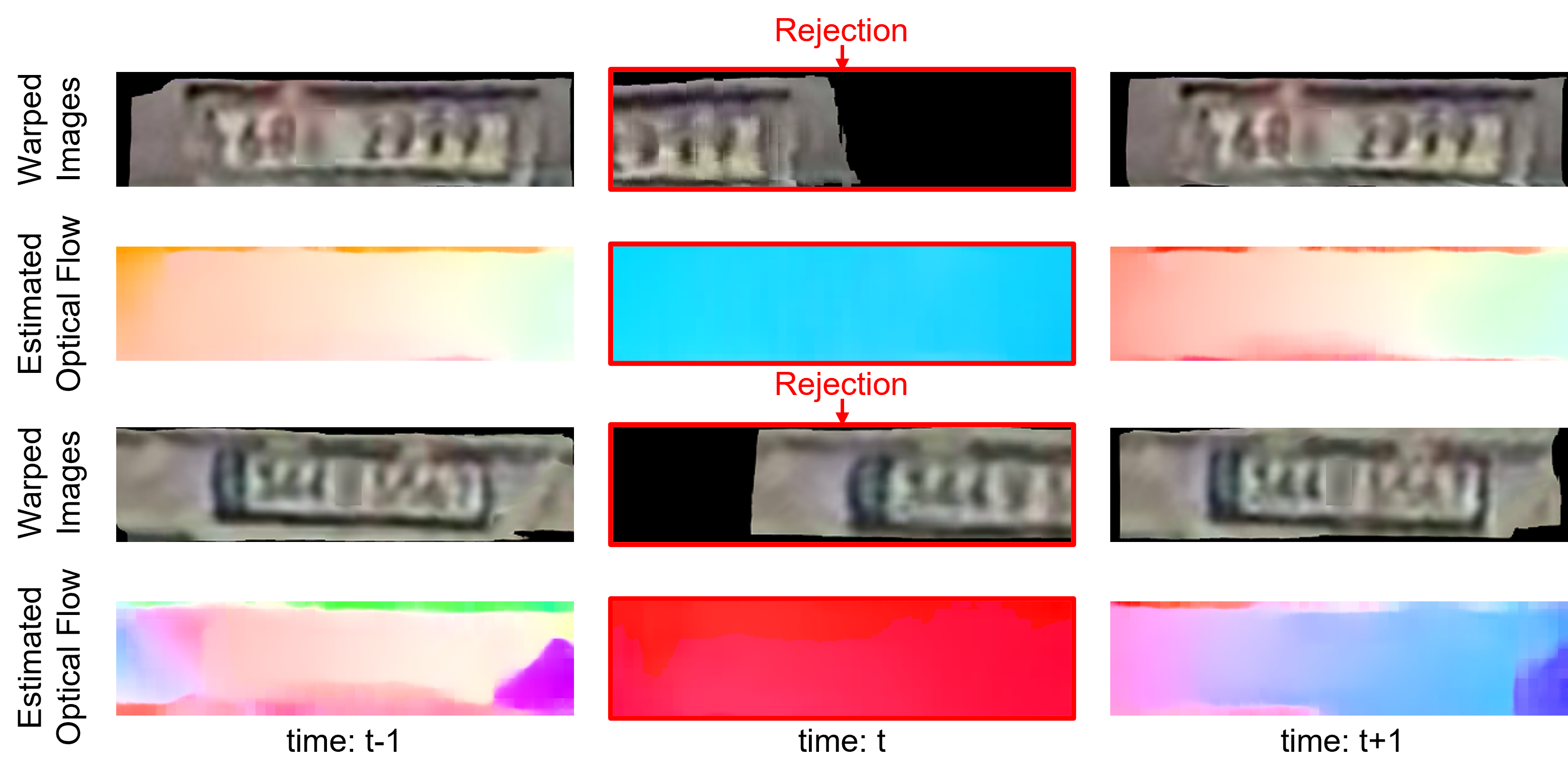}
\caption{Qualitative results of the Temporal Filtering Module in MF-LPR². Our module detects and rejects completely misestimated optical flows, improving LPR accuracy by 1.27\%p. This enhances the inter-frame consistency required for recognition tasks. The solid blue and solid red colors represent misestimated directions, with substantial biases to the left and right, respectively.}
\label{fig:qualitative_temporal}
\end{figure}

\begin{figure}[ht]
\centering
\includegraphics[width=6in]{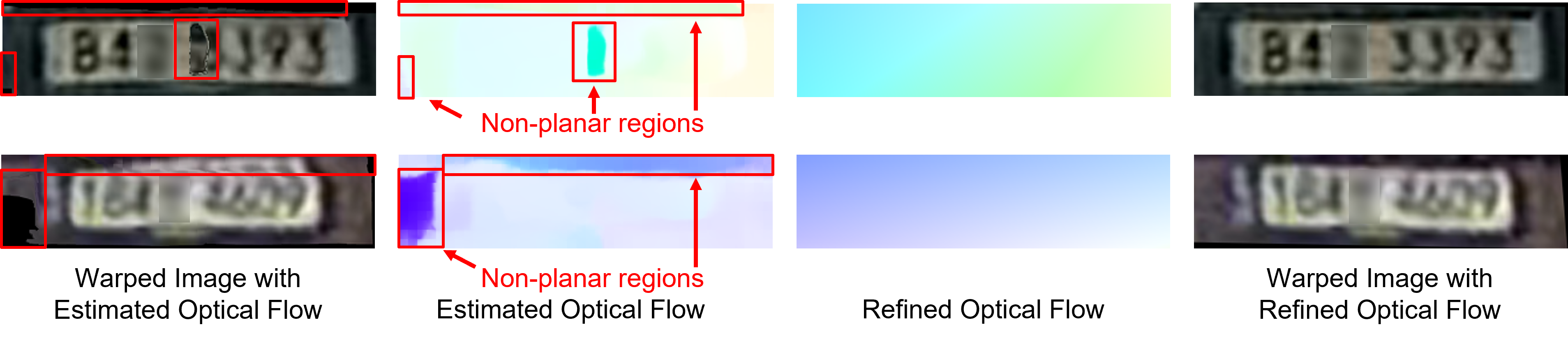}
\caption{Qualitative results of the Spatial Refinement Module in MF-LPR². Our module effectively refines erroneous optical flows by removing non-planar regions, achieving an additional improvement of 10.47\%p in LPR accuracy.}
\label{fig:qualitative_spatial}
\end{figure}

\subsubsection{Temporal Filtering and Spatial Refinement Modules}
\label{subsec:filtering_and_refinement}
To better understand the contributions of the temporal filtering and spatial refinement modules to the overall performance of MF-LPR², we conducted a controlled experiment by developing two model variants and measuring their performance. The first variant only includes the optical flow estimation step and excludes both the temporal filtering and spatial refinement modules. The second variant includes the temporal filtering module but excludes the spatial refinement module. These variants allow us to dissect the impact of each module on key metrics, such as PSNR, SSIM, and License Plate Recognition (LPR) accuracy.

Table \ref{table:module_effectiveness} presents the evaluation results of the three model configurations. Adding the temporal filtering module resulted in improvements in PSNR (from 16.3114 to 16.3240), SSIM (from 0.3468 to 0.3480) and increased LPR accuracy by 1.27\%p (i.e, 1.27 percentage points, from 74.70\% to 75.97\%). Moreover, incorporating both the temporal filtering and spatial refinement modules significantly improved performance across all metrics, achieving the highest LPR accuracy of 86.44\%, an increase of 11.74\%p compared to the baseline model without the modules. This result demonstrates the complementary nature of the two modules, where temporal filtering enhances inter-frame consistency and spatial refinement fine-tunes localized errors, leading to substantial performance gains.

To further validate the effectiveness of each module, we provide qualitative visualizations. Figure \ref{fig:qualitative_temporal} illustrates how the temporal filtering module reduces misaligned flows by rejecting poorly estimated optical flows, while Figure \ref{fig:qualitative_spatial} highlights the spatial refinement module's ability to improve the accuracy of optical flows by refining non-planar regions.

\begin{figure}[th]
\centering
\includegraphics[width=4in]{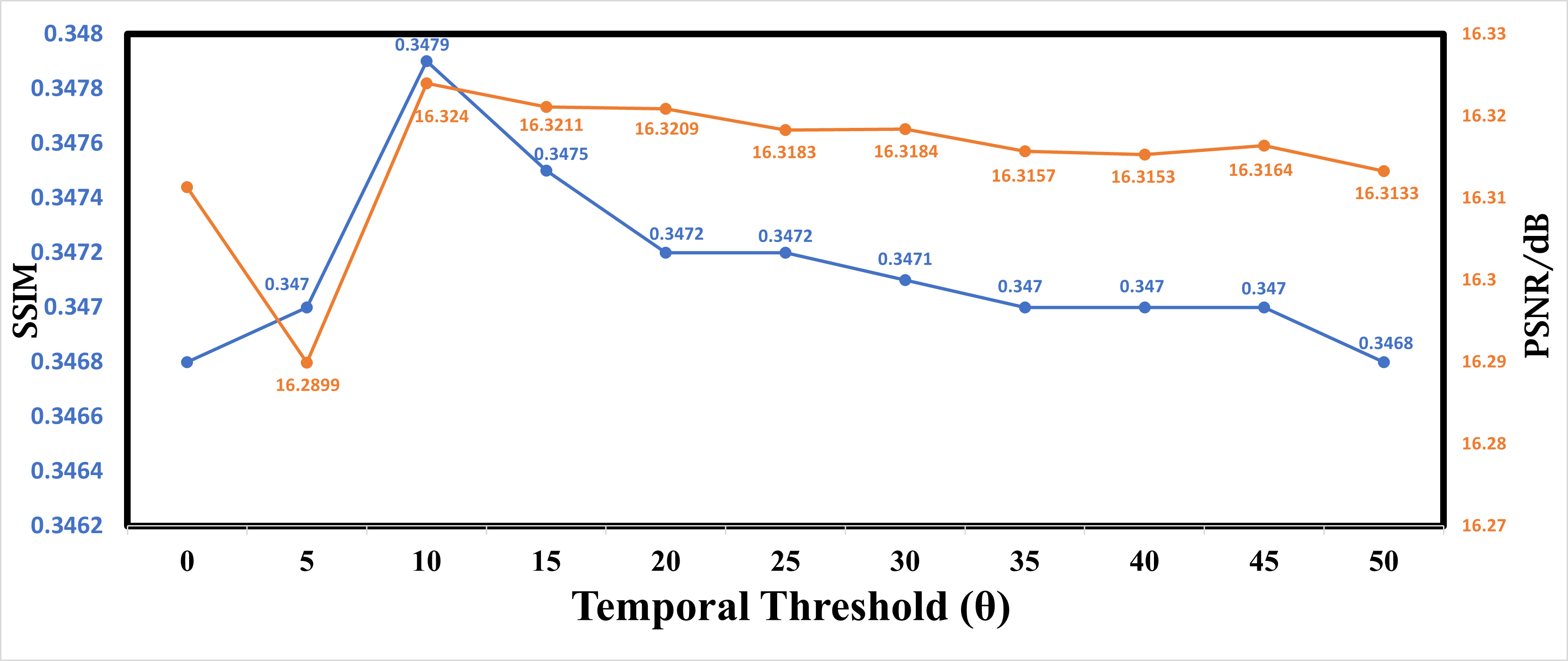}
\caption{The performance of MF-LPR² by the threshold value $\theta_{temp}$ for the temporal filtering module.}
\label{fig:temporal_threshold}
\end{figure}

\begin{figure}[th]
\centering
\includegraphics[width=4in]{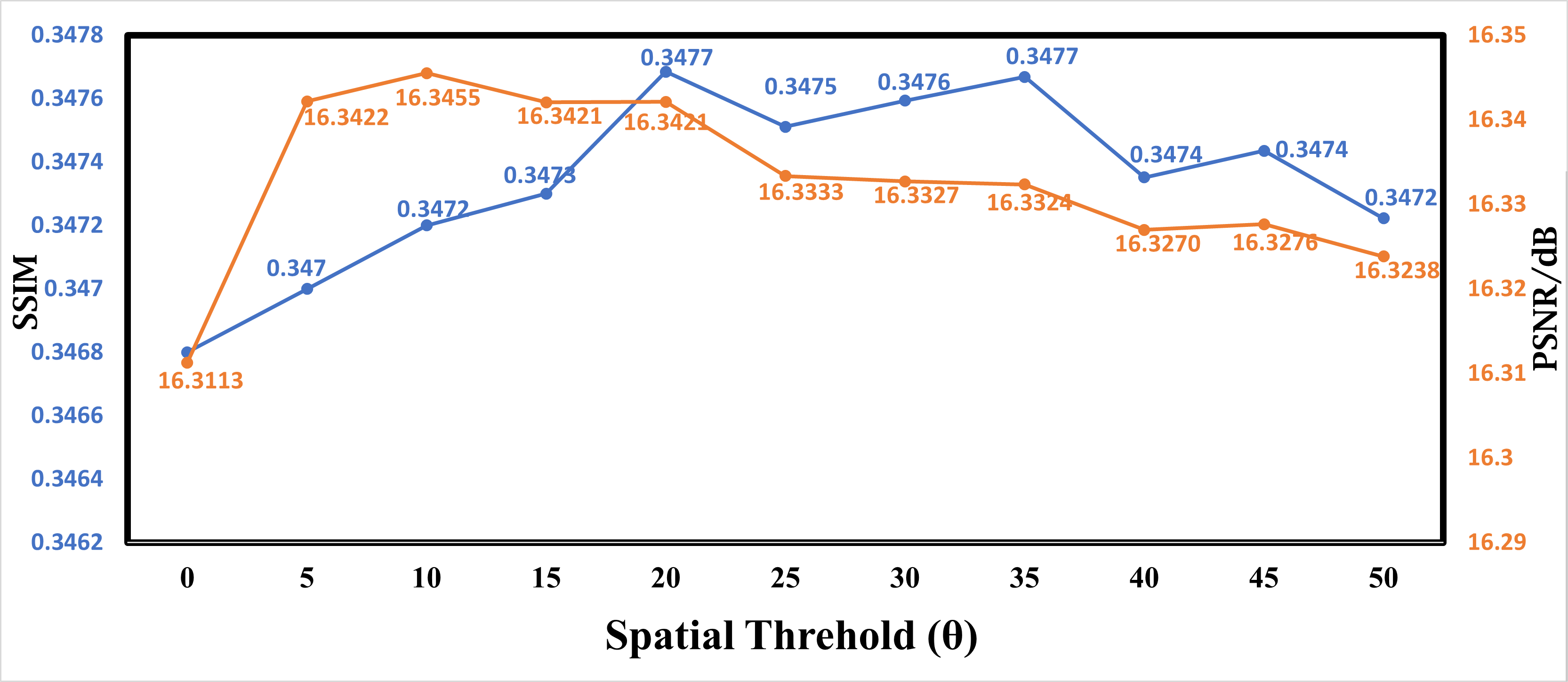}
\caption{The performance of MF-LPR² by the threshold value $\theta_{spatial}$ for the spatial refinement module.}
\label{fig:spatial_threshold}
\end{figure}

\subsubsection{Threshold Values for Temporal Filtering and Spatial Refinement}
To find the optimal values for $\theta_{temp}$ and $\theta_{spatial}$, we measured SSIM and PSNR by changing the threshold values. Fig. \ref{fig:temporal_threshold} and Fig. \ref{fig:spatial_threshold} present the results of ablation studies on $\theta_{temp}$ and $\theta_{spatial}$, respectively.
In Fig. \ref{fig:temporal_threshold}, the highest SSIM value of 0.3479 and PSNR value of 16.324 are observed at a temporal threshold of 10. Beyond this threshold value, both SSIM and PSNR values gradually decrease. The reason of the declination is that, as the temporal threshold increases, fewer images are filtered, leaving more optical flow errors. These errors negatively impact the structural similarity and image quality, thereby causing a reduction in the SSIM and PSNR metrics.
In Fig. \ref{fig:spatial_threshold}, the highest SSIM and PNSR values are observed with different  $\theta_{spatial}$ values. The highest SSIM value of 0.3477 is observed at threshold values of 20 and 35, while the highest PSNR value of 16.3455 is observed at a threshold of 10. Setting $\theta_{spatial}=20$ achieves a good balance, yielding high values in both SSIM and PSNR.

\section{Conclusion}
In this paper, we proposed MF-LPR², a novel framework for multi-frame license plate image restoration and recognition. MF-LPR² effectively addresses the image degradation issues frequently found in low-quality dash cam images, such as insufficient resolution, motion blur, and light glare, by aggregating complementary information from neighboring frames. Unlike most deep-learning-based super-resolution methods that synthesize a new high-quality image, MF-LPR² does not distort the content of the input image or produce spurious artifacts, thereby preserving the evidential value of the original input.
In experiments, MF-LPR² outperformed eight recently developed image/video restoration models and three license plate image restoration and recognition models, by exhibiting a significantly improved character recognition rate. These results suggest that our work introduces meaningful improvements that are useful in various application fields, such as traffic law enforcement, crime investigation, and surveillance.

One limitation of MF-LPR² is that it produces output images solely based on the input images. Referring to contextual information, e.g., utilizing the feedback from the recognizer or leveraging the special characteristics of license plate images, may further improve its performance. These topics can be explored in future work.

\section{Acknowledgments}
\noindent This work was supported by 
NC\& Co., Ltd.,
National Program for Excellence in SW supervised by the Institute of Information\&Communications Technology Planning\&Evaluation (IITP, Korea) supported by Ministry of Science and ICT(MSIT, Korea) in 2023 (2023-0-00055),
and Artificial intelligence industrial convergence cluster development project funded by the Ministry of Science and ICT(MSIT, Korea)\&Gwangju Metropolitan City.

\printcredits
\bibliographystyle{cas-model2-names}

\end{document}